\documentclass[letterpaper]{article} 
\usepackage{aaai23}  
\usepackage{times}  
\usepackage{helvet}  
\usepackage{courier}  
\usepackage[hyphens]{url}  
\usepackage{graphicx} 
\usepackage{natbib}  
\usepackage{caption} 
\usepackage{algorithm}
\usepackage{algorithmic}

\usepackage{amsmath}
\usepackage{amsfonts,amssymb}
\usepackage{multirow}
\usepackage{makecell}
\usepackage{newfloat}
\usepackage{listings}
\DeclareCaptionStyle{ruled}{labelfont=normalfont,labelsep=colon,strut=off} 
\floatstyle{ruled}
\newfloat{listing}{tb}{lst}{}
\floatname{listing}{Listing}
\title{Learning Polysemantic Spoof Trace: A Multi-Modal \\ Disentanglement Network for Face Anti-spoofing}
\author{
    Kaicheng Li\textsuperscript{\rm 1, \rm2},
    Hongyu Yang\textsuperscript{\rm 3}\thanks{Corresponding author.},
    Binghui Chen,
    Pengyu Li,
    Biao Wang,
    Di Huang\textsuperscript{\rm 1, \rm2, \rm4}
}
\affiliations{
    \textsuperscript{\rm 1}State Key Laboratory of Software Development Environment, Beihang University, China\\
    \textsuperscript{\rm 2}School of Computer Science and Engineering, Beihang University, China\\
    \textsuperscript{\rm 3}Institute of Artificial Intelligence, Beihang University, China\\
    \textsuperscript{\rm 4}Hangzhou Innovation Institute, Beihang University, China\\

    \{likaicheng, hongyuyang, dhuang\}@buaa.edu.cn,
        chenbinghui@bupt.edu.cn,
        lipengyu007@gmail.com,
        wangbiao225@foxmail.com
}

\usepackage{bibentry}

\begin{document}

\maketitle
\begin{abstract}

Along with the widespread use of face recognition systems, their vulnerability has become highlighted. While existing face anti-spoofing methods can be generalized between attack types, generic solutions are still challenging due to the diversity of spoof characteristics. Recently, the spoof trace disentanglement framework has shown great potential for coping with both seen and unseen spoof scenarios, but the performance is largely restricted by the single-modal input. This paper focuses on this issue and presents a multi-modal disentanglement model which targetedly learns polysemantic spoof traces for more accurate and robust generic attack detection. In particular, based on the adversarial learning mechanism, a two-stream disentangling network is designed to estimate spoof patterns from the RGB and depth inputs, respectively. In this case, it captures complementary spoofing clues inhering in different attacks. Furthermore, a fusion module is exploited, which recalibrates both representations at multiple stages to promote the disentanglement in each individual modality. It then performs cross-modality aggregation to deliver a more comprehensive spoof trace representation for prediction. Extensive evaluations are conducted on multiple benchmarks, demonstrating that learning polysemantic spoof traces favorably contributes to anti-spoofing with more perceptible and interpretable results.

\end{abstract}

\section{Introduction}


Face Recognition (FR) has been universally accepted and employed in our daily life, involving a diversity of applications, such as mobile payment and access control. But along with its fast popularization, face Spoofing Attacks (SAs, also known as presentation attacks) have become an increasingly critical issue. Generally, the attackers can grab faces of target individuals with cameras or from social networks, which are then counterfeited as reprinted photos, replayed videos or 3D masks for invasion. FR systems encounter enormous threats when attackers masquerade as others. Correspondingly, Face Anti-Spoofing (FAS) is essential.

In the literature, an overwhelming majority of the studies detect SAs with RGB images or video frames, where they mainly perform analysis on the texture difference between the genuine faces and the spoofing ones, \textit{e.g.,} color distortions, unnatural specular highlights, and Moir\'e patterns. The hand-crafted features, such as LBP \cite{LBP2016,freitas2012lbp}, HOG \cite{HOG}, and motion patterns \cite{move1,move2}, are utilized to highlight distinguishable clues in early attempts. In recent years, deep models have been dominating this community. Diverse network structures, including Convolutional Neural Networks (CNNs) \cite{CNNinit}, Recurrent Neural Networks (RNNs) \cite{RNN-LSTM}, and Transformer \cite{transformer} are adopted to learn spoof-relevant spatial and/or temporal features. Additionally, a series of auxiliary tasks \cite{depthrPPG, PatchDepth, reflex1} are presented for improved performance. Unfortunately, these methods are 
elaborately designed for a certain type of attack, 
making them not so competent for more complicated cases, \textit{e.g.}, a wide variety of spoof types co-existing
\cite{liu2019deep}. 
On the other side, generative adversarial deep learning is investigated, and several disentanglement networks are built, which aim to explicitly extract the most fundamental cues in spoofing, \textit{i.e.}, spoof traces, from input faces \cite{STDN, PHYSTDN}. They exhibit great potential for coping with both seen and unseen spoof scenarios. Besides, the disentangled spoof traces also reinforce the interpretability of deep FAS models.

Indeed, it is always desirable that sufficient spoofing clues can be captured by RGB cameras. However, due to the intrinsic limitations of the visible spectrum, it is intractable to detect certain types of attacks only by texture analysis, and sophisticated spoofing techniques, \textit{e.g.,} high-quality replayed videos and high-fidelity 3D masks, make this issue even severer. Considering that multi-modal data convey richer information of faces, a number of efforts are made to reinforce FAS by capturing more comprehensive features, where hyper-spectral \cite{multispec}, thermal \cite{WMCA}, near-infrared \cite{CeFA} and depth \cite{CASIA-SURF} images are exploited. With more comprehensive description, these methods prove superior to the ones based on single-modal inputs \cite{CMFL,OCCL,WMCA}. 
This fact suggests a promising alternative to guard FR systems from spoofing attacks in particular for more practical scenarios, that a versatile FAS model can be constructed by leveraging disentanglement learning as well as multi-modal analysis in a unified network. 
Nevertheless, current multi-modal studies primarily formulate FAS as  a classification problem, where spoof-irrelevant cues are inevitably harvested, and this tends to limit the generalization capability. 

In this study, we propose a novel approach to FAS, which integrates the advantages of disentanglement networks and multi-modal fusion and collaboratively learns polysemantic spoofing characteristics under a generative framework. First, we choose RGB-D modalities as input, since they provide crucial textures and geometries and the devices are relatively affordable and easy to deploy. Then, based on the adversarial learning mechanism, a two-stream disentanglement model is designed to estimate spoof patterns from the RGB and depth modalities, respectively. In this case, it captures complementary spoofing clues inhering in different attacks. Furthermore, instead of directly concatenating the representations learned in the two modalities, an attention-based fusion module is introduced, which recalibrates heterogeneous representations at multiple stages to assist feature disentanglement in each individual modality. It finally performs cross-modality aggregation for a more powerful spoof trace representation. At the decision phase, the spoof patterns decomposed from both the modalities are fed into a vanilla classification network to predict the attack probability score.


The main contributions include: (i) the first multi-modal disentanglement network for FAS, which effectively improves the  accuracy and robustness by learning polysemantic spoof traces;
(ii) a feature fusion module, which mines helpful clues across modalities to facilitate the disentanglement; and (iii) the state-of-the-art spoofing detection performance on several publicly available benchmarks under both the seen and unseen protocols.



\section{Related Work}
\subsection{Conventional Face Anti-spoofing}
The early face anti-spoofing approaches mostly utilize hand-crafted features
to extract potential color and texture differences caused by spoofing attacks \cite{LBP2016,HOG,SIFT}. Several studies construct dynamic features to emphasize facial actions like blinking \cite{eyeblink} and micro movements \cite{move1,move2}. In the context of deep learning, CNN- and RNN-based methods are introduced \cite{CNNinit,RNN-LSTM} and a series of auxiliary tasks, including recovering depth \cite{depth2,PatchDepth}, reflection \cite{reflex1,reflex2}, and rPPG signals \cite{depthrPPG}, are considered as constraints to capture fine-grained textural differences. Besides, specific pre-processing and pixel-level supervision are developed to enhance local representations \cite{patch1,DeepPixBis,pix2}. 
However, these solutions heavily deteriorate in practical scenarios  where unseen spoofing attacks appear. To solve this problem, a few generalizable methods are designed to mitigate the image distribution shift by performing domain adaptation \cite{DA_MMD_2018,IJCB19UDA,TIFS20UDA,MADDG,SSDG}. 
Meanwhile, meta-learning based ones are explored \cite{Meta2,meta3,meta4}. Despite that gains are continuously reported, the methods above are largely limited by the single RGB modality.

\subsection{Multi-Model Face Anti-spoofing}
\cite{WangYan} jointly employs the texture features and the  geometric ones extracted from reconstructed depth maps to perform FAS on 3D masks. \cite{SURF_dataset} proposes a multi-branch network with a squeeze and excitation fusion model to combine RGB, depth, and near-infrared features. \cite{SD_SE} further introduces a multi-level fusion branch for improved results. \cite{ChenSong} presents MM-DFN to model multi-modal dynamic features. \cite{CBAM_multi} adopts CBAM \cite{CBAM} to build modality-specific liveness features. \cite{faceBagNet} takes patch-level input and designs a modal feature erasing operation to prevent over-fitting. \cite{CMFL} makes use of a cross-modal focal loss to adjust the contribution of each modality while suppressing the impact of uncertain samples. Additionally, input-level and decision-level fusion schemes are considered in \cite{inputfusion1,inputfusion2}. However, these multi-modal methods have the difficulty in generalizing to unseen spoofing attacks and along with the conventional ones, their results are hard to interpret.  



\begin{figure*}[t]
\centering
\includegraphics[width=0.94\textwidth]{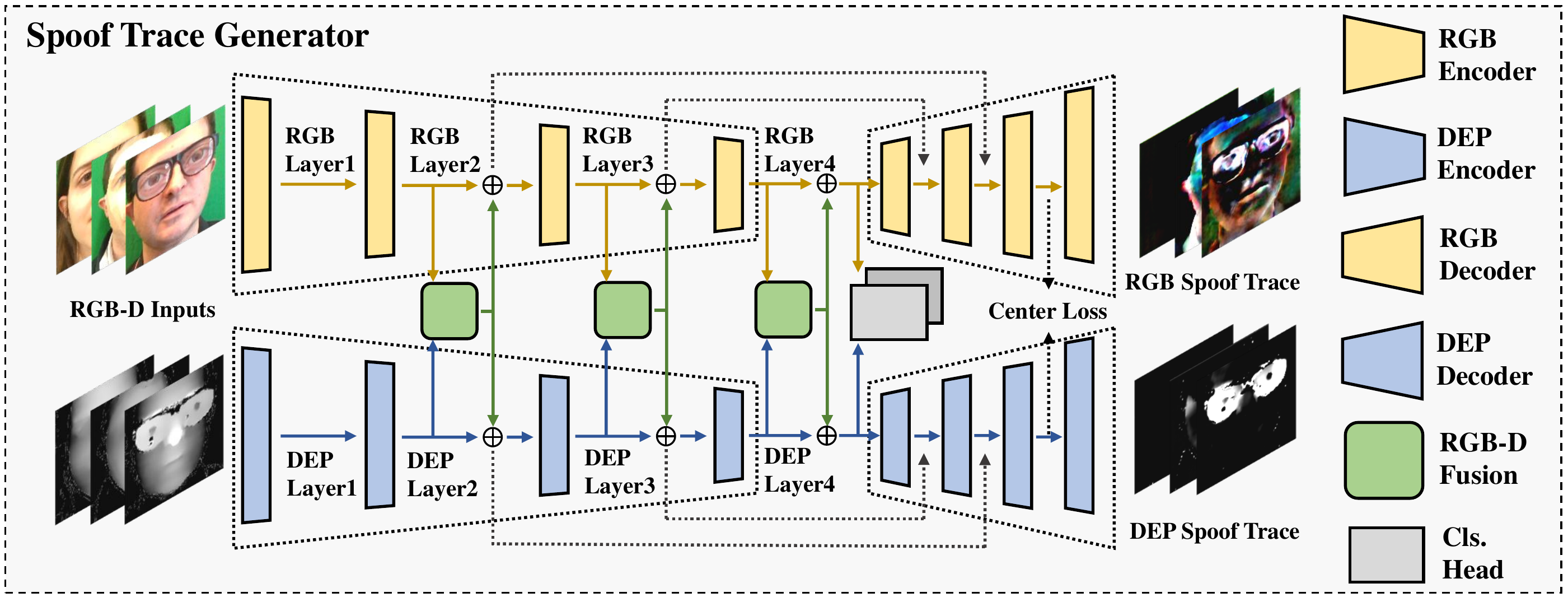} 
\caption{Architecture of the spoof trace generator. It consists of two sub-networks to explicitly disentangle polysemantic spoof traces from the RGB-D inputs. The intermediate features are recalibrated by a cross-modality fusion module at multi-scales.}

\label{myfig1}

\end{figure*}

\subsection{Generative Face Anti-spoofing} 
A few disentanglement frameworks are proposed to explicitly decompose spoof patterns from input faces. \cite{DeSpoofing} rephrases FAS as a noise modeling problem and establishes an encoder-decoder structure to estimate spoof noise. \cite{STDN,PHYSTDN, xu2021identity} then construct a series of spoof trace elements by adversarial learning. \cite{spoofcue} applies the classification and metric loss to generate spoof cues in an anomaly detection manner. \cite{xu2021identity} exploits identity information to constrain the estimation of the live face components and generates spoof noise for 2D attacks. Apart from directly generating spoof patterns, several studies use the latent code to disentangle spoof and content features \cite{latent1,latent2, latent3}.
Nevertheless, the previous methods mostly focus on dealing with 2D attacks in the RGB modality. In this study, we leverage the disentanglement mechanism and substantially extend it to the multi-modal space, where we specially investigate the interactions between modalities and present a network to learn polysemantic spoof traces in a collaborative manner. As in \cite{CMFL}, RGB-D data are used in this study, but our model can be conveniently adapted to more modalities.

 
\section{Method}
Our model performs FAS on RGB-D images, which are biased towards textural and geometric spoof clues, respectively. 
As illustrated in Figure \ref{myfig1}, the model is implemented as a generative network, where the two branches explicitly disentangle polysemantic spoof traces from different modalities. To collaboratively learn spoof representations, the intermediate features are recalibrated by a cross-modality fusion module at multi-scales, which not only assists to propagate complementary information between modalities but also helps to preserve modality-specific information. A series of training losses are exploited to guide the training process for improved accuracy and robustness.
Both types of the estimated spoof traces are fed into a classification network to make final decision.

\subsection{Spoof Trace Modeling}

Motivated by \cite{STDN, xu2021identity}, we regard spoof trace disentangling as a generative problem, where a face can be decomposed into a spoof trace, \textit{i.e.}, the detectable artifact introduced by attacking, and its live counterpart.
We uniformly exploit a denoise model \cite{DeSpoofing}, ensuring that it can be conveniently generalized.
For any single modality, an input face image $I$ can be formulated as:
\begin{equation}
\begin{aligned}
I=\hat{I}+T
\end{aligned}
\label{modeling}
\end{equation}
where $\hat{I}$ refers to the live component and $T$ indicates the spoof trace. Theoretically, the decomposed live components should share the same distribution with the real live samples; therefore we extract a spoof trace by reconstructing its original live part via bidirectional adversarial learning. The spoof trace is expected to fully contain spoof-relevant cues, upon which the attack can be detected.

\subsection{Spoof Trace Generator}
Multiple modalities can reinforce the model. The fine-grained color and texture differences caused by spoof instruments are generally conveyed in RGB images,
while depth maps mainly contain corresponding geometry variations. To identify the respective strengths of the two modalities and integrate their complementary representations, the spoof trace generator is implemented as a two-stream network to targetedly learn polysemantic spoof traces. By using a cross-modal fusion module, the generator spotlights the intermediate modality-specific features at multiple stages and unifies the features into more comprehensive representations to deliver the final spoof traces.


For either branch, the encoder is constructed by the first three blocks of the ResNet-50 model pre-trained on ImageNet. It gradually encodes the input RGB or depth image into a latent space capturing the spoof-relevant characteristics. 
The decomposition is achieved by a CNN-based decoder, which consists of multiple stacked residual blocks and deconvolutional layers, yielding the spoof trace conditioned on the input. In the backbone, skip connections are applied for a high-quality transformation. Besides, the cross-modality fusion module is used. To better guide the learning process of the spoof trace generator, an intermediate classification head is also built to encourage the encoder to focus on the spoof-relevant information. It takes the bottleneck feature of the branch as input and outputs a classification score.

\subsection{Cross-Modality Feature Fusion} 
\begin{figure}[t]
\centering
\includegraphics[width=0.48\textwidth]{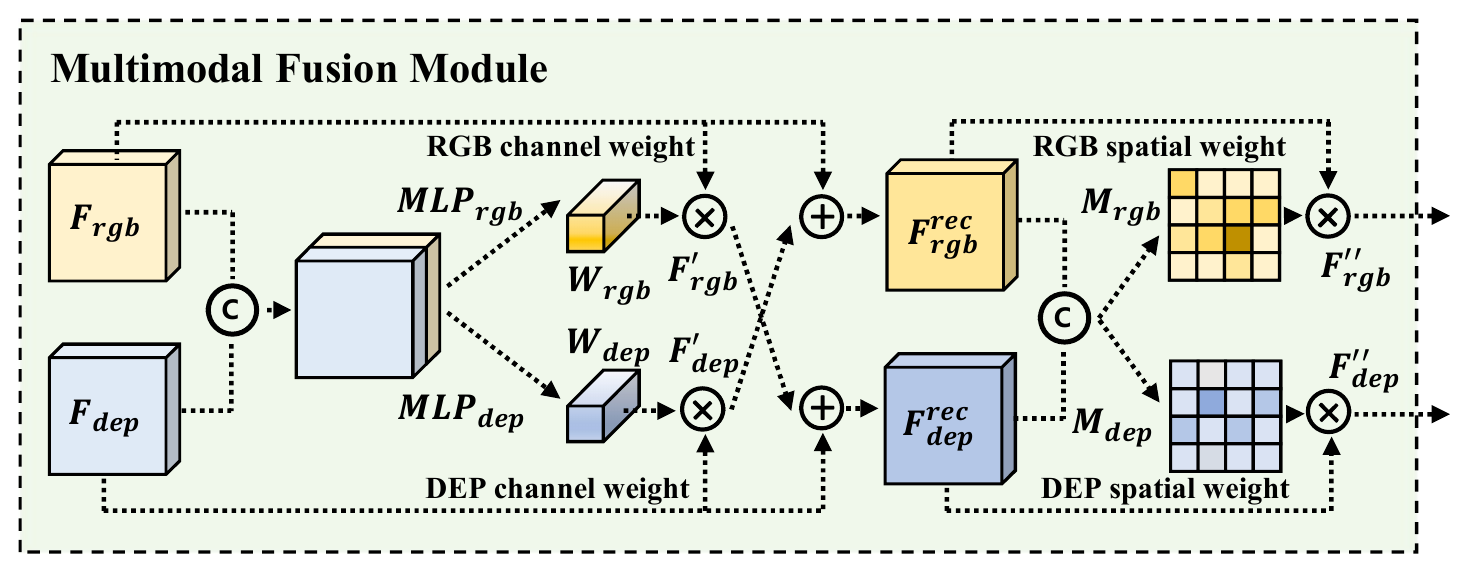} 
\caption{Illustration of the cross-modality fusion module.}
\label{figattention}
\end{figure}

Multi-modal data naturally complements each other, but simply assembling heterogeneous information possibly leads to inferior results. 
Our fusion strategy is inspired by SA-Gate \cite{SAGATE}, which first conducts Feature Separation (FS) to select the most informative feature maps of each modality via a cross-modality channel attention, and then conducts Feature Aggregation (FA) by adding the cross-modality features with a spatial-wise gate. We take the advantage of SA-Gate in leveraging both channel-wise and spatial-wise correlations. Unlike the original SA-Gate which produces a single fused representation, we put a higher emphasis on the specificity of the feature in each modality, \textit{i.e.}, the RGB and depth information is individually enhanced. In this way, the two kinds of features are then propagated to the next stage of the two-branch generator and their respective spoof traces can be obtained. The fusion module is shown in Figure \ref{figattention}. 

We take the RGB modality as an example. Let $F_{rgb}$ and $F_{dep}$ denote the learned RGB and  depth features. The two types of features are first concatenated, and to filter the most important channels, an MLP is trained to calculate the channel attention between the modalities, followed by a sigmoid function rescaling the weights:
\begin{equation}
\begin{aligned}
W_{rgb} =& \sigma(MLP_{rgb}(F_{rgb}||F_{dep}))\\
\end{aligned} 
\label{W_rgb}
\end{equation}
A filtered RGB representation, denoted as $F^{'}_{rgb}$, can be obtained via a channel-wise multiplication between the input RGB feature and the cross-modality gate, written as:
\begin{equation}
\begin{aligned}
F^{'}_{rgb} =& W_{rgb}\otimes{F_{rgb}} \\
\end{aligned} 
\label{F_rgb}
\end{equation}
where $\otimes$ denotes channel-wise multiplication. 
With the enhanced RGB feature, its depth counterpart can be further recalibrated by borrowing such complementary information from the RGB branch:
\begin{equation}
\begin{aligned}
F^{rec}_{dep} =& F^{'}_{rgb} + F_{dep}  \\
\end{aligned} 
\label{F_rec}
\end{equation}

Similarly, we have $F^{'}_{dep}$ and $F^{rec}_{rgb}$. The recalibration step is conducted in a symmetric manner so that both modalities can be spotlighted by making them aware of global information. Next, with the refined features, a spatial aggregation step is performed to harvest useful spoof-related context within each modality. In particular, the spatial attention weight is calculated via a $1\times 1$ convolutional layer followed by softmax, which enforces the model to focus on the regions introduced by attacks. Finally, the reinforced feature can be formulated as:  
\begin{equation}
\begin{aligned}
F^{''}_{rgb} =& F^{rec}_{rgb} \odot M_{rgb}(F^{rec}_{rgb}||F^{rec}_{dep})\\
F^{''}_{dep} =& F^{rec}_{dep} \odot M_{dep}(F^{rec}_{rgb}||F^{rec}_{dep})\\
\end{aligned} 
\label{F_rgb_}
\end{equation}
where $M_{rgb}$ and $M_{dep}$ denote the convolutional operations for the RGB and depth modalities, respectively, and $\odot$ is element-wise multiplication. 
To fully integrate the two modalities, cross-modality feature fusion is conducted at multiple layers, which enables the complementary information to propagate between modalities throughout the network. With such a network architecture, the RGB and depth features are learned in a mutually reinforced manner, leading to improved disentanglement performance.

\begin{figure}[t]
\centering
\includegraphics[width=0.48\textwidth]{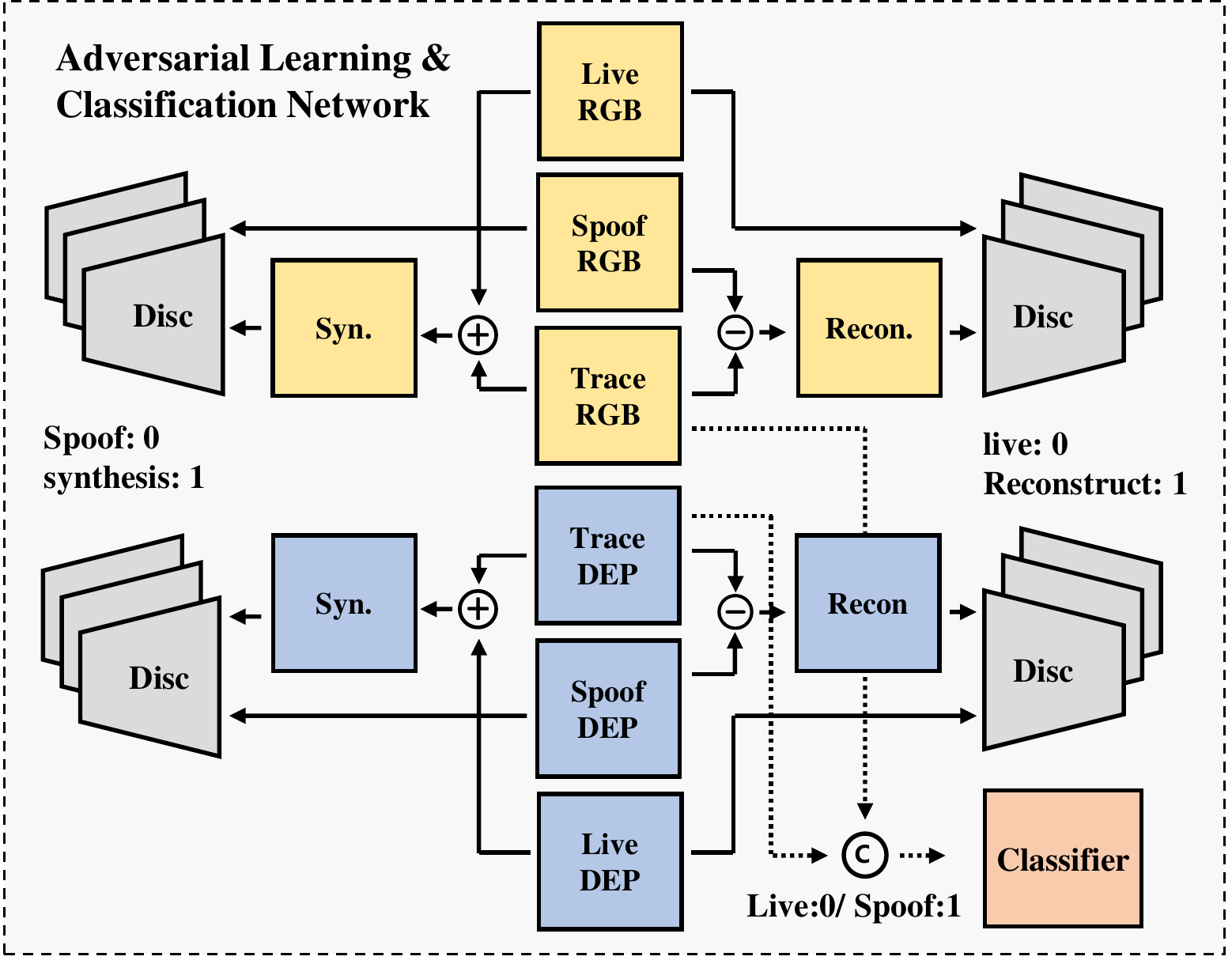} 
\caption{Illustration of the adversarial learning procedure and the final classification network. $Disc$ refers to a set of multi-scale discriminators. 
}
\label{fig_adv_cls}
\end{figure}

\subsection{Training Process and Loss Functions}
\textbf{Adversarial Loss.} Considering that the pixel-level ground-truth labels for spoof patterns are unavailable and the learning process cannot be directly supervised \cite{DeSpoofing}, we adopt a bidirectional adversarial learning mechanism to achieve spoof trace disentanglement, as illustrated in Figure \ref{fig_adv_cls}. Specifically, we leverage the following observations: 
(1) The live components disentangled from the input images and the real face samples are expected to share the same data distribution, and we thus adopt the \textbf{general adversarial learning} to encourage the separated live counterparts to be indistinguishable from the real faces. (2) The decomposed spoof patterns can be used to synthesize new spoof images for enhanced disentanglement. For simplicity, we add the obtained spoof pattern to a randomly selected live image to render a synthesized spoof face. In an ideal situation, the synthesized spoof images should share the same distribution with the actual spoof samples. Therefore, the \textbf{symmetrical adversarial loss} is calculated.

Motivated by \cite{STDN}, multi-scale discriminators are used to estimate the features and 
the PatchGAN structure \cite{PatchGan} is adopted, as finer-scale supervision is more suitable for tackling detailed spoof traces. The discriminator is implemented with 7 convolutional layers and 3 down-sampling layers. The adversarial loss for the generator can be formulated as:
\begin{equation}
\begin{aligned}
L_{G}=& \mathbb{E}_{i\in{L\cup S}}{\lbrack{\Vert D_{L} (\hat{I_{i}})\Vert}_{2} \rbrack} +\mathbb{E}_{i\in{L},j\in{S}}{\lbrack{\Vert D_{S} (I_{i}+T_{j})\Vert}_{2} \rbrack} 
\end{aligned}
\label{Gadvloss}
\end{equation}
where $D$ is the multi-resolution discriminator set, and $L$ and $S$ denote the set of live and spoof faces, respectively. $T_j$ is the multi-modal spoof trace disentangled from a spoof sample $I_j$, and $I_i$ denotes a randomly selected live sample. The discriminators of the two modalities are trained separately. The adversarial loss pushes them to distinguish between the real live samples and the reconstructed ones as well as the actual spoof samples and the synthesized spoof ones, written as: 


\begin{equation}
\begin{aligned}
L_{D}=& \mathbb{E}_{i\in{L \cup S}}{\lbrack{\Vert D_{L}(\hat{I_i})-1\Vert}_{2} \rbrack} + \mathbb{E}_{i\in{L}}{\lbrack{\Vert D_{L}(I_i)\Vert}_{2} \rbrack} +\\
&\mathbb{E}_{i\in{L},j\in{S}}{\lbrack{\Vert D_{S} (I_{i}+T_{j})-1\Vert}_{2} \rbrack} +\mathbb{E}_{i\in{S}}{\lbrack{\Vert D_{S} (I_{i})\Vert}_{2} \rbrack}
\end{aligned}
\label{Dadvloss}
\end{equation}

 \noindent \textbf{Identity Consistency Loss.} Only using the adversarial loss would cause the decomposed live components to be biased towards the average, and the spoof-irrelevant information, \textit{e.g.}, identity, would be squeezed into the generated spoof trace, especially when the training samples are not so sufficient. 
 To avoid this, an identity consistency loss is utilized to constrain the reconstructed and original input face so that the identity property is preserved. We adopt a lightweight ArcFace model \cite{arcface} based on ResNet-18 as the face recognizer, which measures the identity similarity between the input and reconstructed samples:

\begin{equation}
\begin{aligned}
L_{id}= \mathbb{E}_{i\in{L \cup S}}{\lbrack{\Vert f_{Arcface}(I_i)-f_{Arcface}(\hat{I_i})\Vert}_{2} \rbrack}
\end{aligned}
\label{idloss}
\end{equation}
where $f_{Arcface}$ is the pertained face recognition model.

 \noindent \textbf{Intensity Loss.} Since the real live sample and its live counterpart should be the same, an intensity regularizer is exploited to constrain its spoof trace to be zeros. We also regularize the intensity of the spoof patterns to avoid outliers:
\begin{equation}
\begin{aligned}
L_{intensity}= &\mathbb{E}_{i\in{L}}{\lbrack{\Vert T_{i}\Vert}_{1}\rbrack} +  \lambda_{t}\mathbb{E}_{i\in{S}}{\lbrack{\Vert T_{i}\Vert}_{1} \rbrack}
\end{aligned}
\label{regloss}
\end{equation}

\noindent \textbf{Center Loss.}  The center loss \cite{centerloss} is adopted to optimize the intermediate feature distribution, targeting improved generalization capability to unknown attacks. For each modality, the feature output by the 3rd up-sampling layer of the decoder is used to compute the loss, mainly because this layer is after all the skip-connections thus contains more distinguishable FAS information. The center loss is individually calculated on each modality:
\begin{equation}
\begin{aligned}
L_{center}= &\frac{1}{2}\sum\limits^2\limits_1{{\Vert x_{i}-c_{y_i}\Vert}^{2}_{2}}
\end{aligned}
\label{centerloss}
\end{equation}
where $c_{y_i}$ denotes the feature 
centers of the live and spoof samples, respectively, and $y_i$ refers to the ground-truth label which is set to $0$ for the live samples and $1$ for the spoofs. The calculation of the updated $c_{y_i}$ remains the same as in \cite{centerloss}.


\noindent \textbf{Classification Loss.} The classic cross entropy loss is calculated on the intermediate heads of the two encoders:

\begin{equation}
\begin{aligned}
p_{i} = &\sigma(\psi(E(I_i)))\\
L_{e} = &\sum\limits^n\limits_{i=1}(y_{i}log(p_i)+(1-y_{i})log(1-p_{i}))
\end{aligned}
\label{ESRloss}
\end{equation}
where $E$ is the feature encoder and $\psi$ denotes the classification head of each branch.


\begin{table}[t]
\centering
\begin{tabular}{c|c|c}
    \hline
     Modality & Method & ACER \\
    \hline
    \multirow{3}*{\thead{RGB-D\\IR\\ Thermal}}& MCCNN-BCE & \textbf{0.2}\\
    & MCCNN-OCCL-GMM & 0.4\\
    & Conv-MLP & 0.9\\
    \hline
     \multirow{7}*{RGB-D}& MC-PixBiS & 1.8\\
     & MCCNN-OCCL-GMM & 3.3\\
     & MC-ResNetDLAS & 4.2\\
     & Conv-MLP & 6.0 \\
     & TTN-T-NHF & 0.8\\ 
     & TTN-S-NHF & 0.3\\ 
    
     \cline{2-3}
     &Ours & \textbf{0.27}\\
    \hline
\end{tabular}
\caption{The evaluation on WMCA, GrandTest protocol.}
\label{grandtest}

\end{table}

\begin{table*}[t]\small
\centering
\begin{tabular}{c|lrrrrrrrr}
\hline
Modality& Method  &FlexibleMask &Replay &FakeHead &Prints &Glasses &PaperMask &RigidMask &Mean ± Std\\
\hline
\multirow{5}*{RGB} & ResNet50& 14.5 &15.7 &38.0 &32.7 &27.3 &20.1 &30.2&25.5±9.0\\
& CDCN & 12.1 & 8.7 &42.7 &30.1 & 11.7 &11.9 & 30.4 &21.1±13.2\\
& Auxiliary(Depth) & 13.2 & 12.5 &47.3 &32.2 &23.7 &13.9 & 40.4 &26.2±14.1\\
& TTN-T-NHF & 15.1 &33.8 & 1.3 &0.4 &40.4 &3.0 &6.0 &14.3±16.4\\
& TTN-S-NHF & \textbf{10.7} &21.9 & 1.3 &\textbf{0.0} &25.4 &\textbf{0.0} & 2.0 &8.8±9.9\\
\hline
\multirow{7}*{RGB-D} & MC-PixBiS& 49.7 &3.7 &0.7 &0.1 &16.0 & 0.2 &3.4 &10.5±16.7\\
& MCCNN-OCCL &22.8 &31.4 &1.9 &30.0 &50.0 &4.8 &18.3 &22.7±15.3\\
& ResNetDLAS &33.3 &38.5 &49.6 &3.8 &41.0 &47.0 &20.6 &33.4±14.9\\
& CMFL & 12.4 &1.0 &2.5 &0.7 &33.5 &1.8 &1.7 &7.6±11.2\\
& TTN-T-NHF & 26.4 &\textbf{0.0} &\textbf{0.0} &\textbf{0.0} &15.9 &1.8 &8.0 &7.4±10.2\\
& TTN-S-NHF & 21.7 &1.7 &1.7 &0.0 &21.3 &0.7 &2.3 &7.1±9.9\\
     \cline{2-10}
& Ours& 19.0 & 0.5 & 2.3 & 0.7 & \textbf{10.0} & 0.7  &\textbf{0.6}  &\textbf{4.8±6.6}\\
\hline

\end{tabular}
\caption{The ACER (\%) results achieved on WMCA, LOO protocol.}
\label{WMCA_LOO}
\end{table*}

During training, each mini-batch involves three training steps: generator step, discriminator step, and consistency supervision step. 
The overall loss of the generator step is:
\begin{equation}
\begin{aligned}
L_{}= &\alpha_{1}L_{G}+\alpha_{2}L_{intensity}+\alpha_{3}L_{e}+\\
&\alpha_{4}L_{id}+\alpha_{5}L_{cls}+\alpha_{6}L_{center}
\end{aligned}
\label{step1loss}
\end{equation}

At the discriminator step, the model is optimized by $\alpha_{7}L_{D}$,
and at the consistency supervision step, the disentangled spoof traces are used to synthesize new spoofs as stated previously. 
The generator is expected to regenerate these spoof patterns:
\begin{equation}
\begin{aligned}
L_{}= &\alpha_{8}\mathbb{E}_{i\in{L},j\in{S}}{\lbrack{\Vert G(I_i+T_j)-T_j\Vert}_{2} \rbrack} 
\end{aligned}
\label{consistencyloss}
\end{equation}
Here, to obtain a better generalization ability, the traces of two different spoof samples are also randomly mixed to increase the variety. 

 \noindent \textbf{Inference.} The final polysemantic spoof trace is obtained by concatenating the decomposed multi-modal spoof traces in channel, which is fed into a vanilla fully-convolutional network $\phi$ to perform classification. 

\section{Experiments}
\subsection{Experimental Settings}
\subsubsection{Databases and Metrics}
We conduct extensive experiments on two benchmarks, \textit{i.e.}, WMCA \cite{WMCA} and CASIA-SURF CeFA \cite{CeFA}. \textbf{WMCA} contains seven kinds of 2D and 3D facial presentation attacks, including print, replay, spoof glass, fake head, silicon mask, paper mask, and rigid mask. Multiple modalities are captured in color, depth, infrared, and thermal images.
We follow the two major testing protocols on WMCA, \textit{i.e.}, \textit{GrandTest} and \textit{Leave-one-out (LOO)}.
\textbf{CASIA-SURF CeFA} consists of both 2D and 3D attacks, which additionally considers 3D rigid masks and silicon masks. The data modalities contain RGB, depth, and infrared images. There are four testing protocols: \textit{cross-ethnicity}, \textit{cross-PAI}, \textit{cross-modality} and \textit{cross-ethnicity \& PAI}. 
The 3D attack subset is included only in the evaluation set. 
In this study, we only focus on the RGB and depth modalities. 
5 frames are randomly selected from each video for training and testing.
We report APCER (\%), BPCER (\%), and ACER (\%), and the threshold is calculated at BPCER = 1\% on the validation set.


\subsubsection{Implementation Details}
Training is launched on a Nvidia Geforce 1080Ti with a batch size of 4 using the Adam optimizer for 40,000 iterations. The initial learning rate is 5e-5. We resize the input face into $256\times256$ pixels. The hyper-parameter $\lambda_{t}$ is set to 1e-4, and $\lbrace\alpha_{1}, \alpha_{2},\alpha_{3}, \alpha_{4}, \alpha_{5}, \alpha_{6}, \alpha_{7}, \alpha_{8} \rbrace$ are set to $\lbrace 0.25, 100, 1, 100, 1, 10, 1, 1\rbrace$. 

\subsection{Quantitative Evaluations}
WMCA and CeFA both contain more modalities besides RGB and depth, and some previous studies use all the modalities. To make fair comparison, we reimplement the representative methods by their official codes with default settings, where RGB-D data are adopted for training and testing. For WMCA, the compared methods include MC-PixBiS \cite{MC_Pixbis}, MCCNN-OCCL-GMM \cite{OCCL}, MC-ResNetDLAS \cite{SD_SE}, CMFL \cite{CMFL}, and TTN \cite{TTN}, which signify the-state-of-art. For CeFA, the counterparts include MC-PixBiS, FaceBagNet \cite{CVPRW19}, CDCN \cite{MCCDCN} and CMFL.


\begin{figure}[t]
\centering
\includegraphics[width=0.42\textwidth]{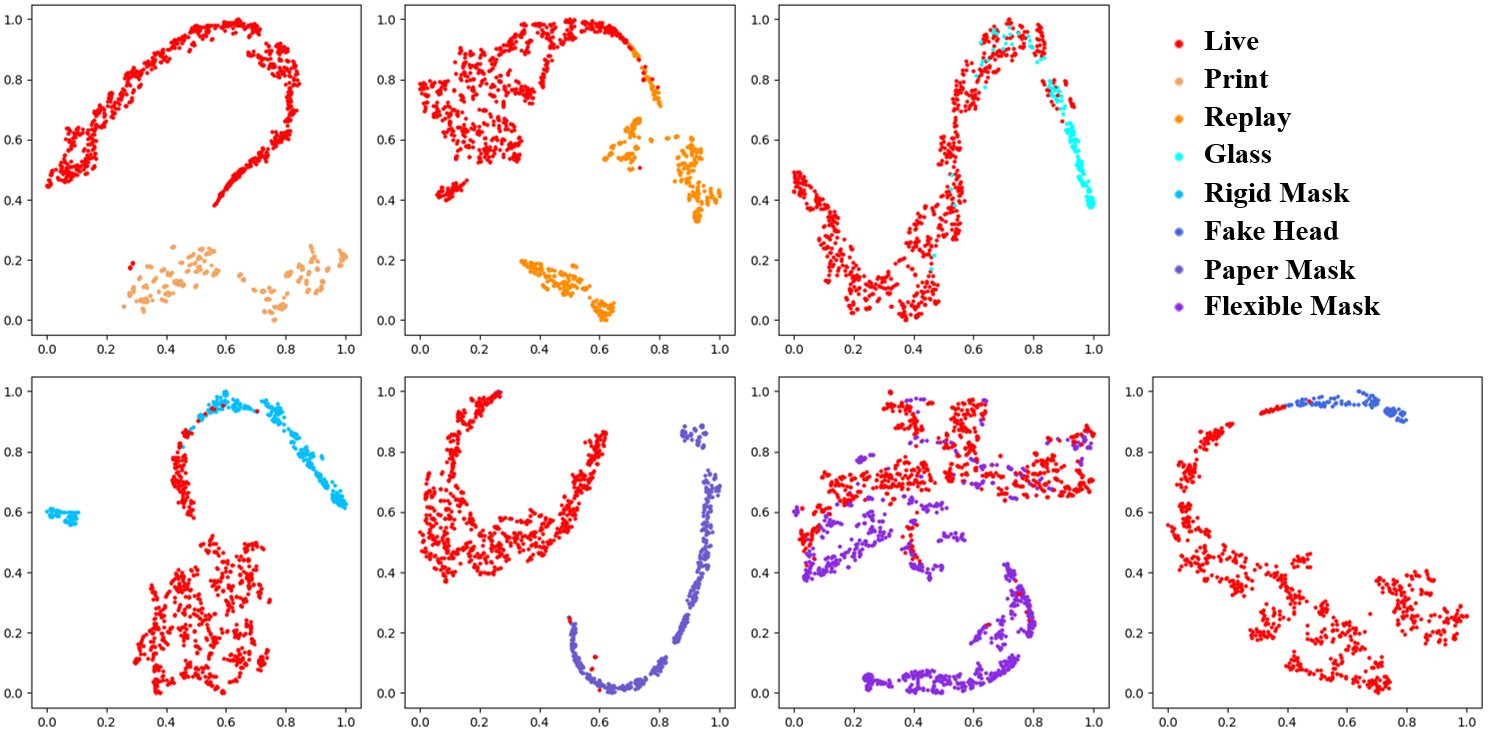} 
\caption{Feature distributions by $t$-SNE on WMCA, LOO protocol.}
\label{TNSE_LOO}

\end{figure}

\subsubsection{WMCA}
\textit{GrandTest Protocol:} This evaluation validates the ability to simultaneously detect multiple spoof types. The results are shown in Table \ref{grandtest}, where the ones of the compared methods are directly quoted from the original papers. Our approach achieves the best performance compared with the others on RGB-D data, reducing the ACER by 0.03\% - 5.73\%. The performance is comparable to or even better than that of the methods which exploit more modalities (including MCCNN-BCE \cite{WMCA}, MCCNN-OCCL-GMM \cite{OCCL}, and Conv-MLP \cite{CONV-MLP}), clearly indicating its effectiveness in complex spoofing scenarios.

\begin{table}[t]\small
\centering
\begin{tabular}{c|c|ccc}
    \hline
    Prot. & Method & ACER & APCER & BPCER \\
    \hline
    \multirow{5}*{1} &MC-PixBiS& 15.1±6.9 &1.4±0.8 & 28.7±14.6 \\
     &FaceBagNet & 17.4±7.9 & 2.1±1.8 & 32.7±17.0   \\
     &CDCN & 6.8±3.5 & \textbf{0.0±0.0} & 13.6±7.0  \\
     &MCFL & 13.5±9.8 & 3.7±4.4 & 23.4±15.6 \\
     &Ours  & \textbf{5.0±1.9} & 2.1±3.5 & \textbf{7.9±5.3} \\

    \hline
    \multirow{5}*{2} &MC-PixBiS & 5.9±7.5 & 10.6±14.9 & 1.2±0.2 \\
     &FaceBagNet & 7.9±10.3 & 15.5±21.0 & 0.9±0.2  \\
     &CDCN     & 1.2±0.6 & \textbf{0.0±0.0} & 2.5±1.3  \\
     &MCFL     & 2.2±2.6 & 3.6±5.1 & \textbf{0.9±0.0} \\
     & Ours     &\textbf{ 0.9±0.2} & 0.2±0.4 & 1.5±0.7 \\

    \hline
    \multirow{5}*{4} &MC-PixBiS& 15.9±1.4 & 19.0±4.8 & 12.8±4.6 \\
     &FaceBagNet & 26.7±9.8 & 37.9±21.0 & 15.4±3.2  \\
     &CDCN & 9.7±6.1 & \textbf{0.5±1.0} & 19.0±12.6  \\
     &MCFL & 15.2±8.6 & 11.6±3.4 & 19.0±21.2 \\
     &Ours  & \textbf{6.2±1.0} & 1.3±0.6 & \textbf{11.1±2.2} \\
    \hline
\end{tabular}
\caption{The evaluation on CeFA, LOO protocol.}
\label{table3}
\end{table}

\begin{table}[t]\small
\centering
\begin{tabular}{c|ccc}
    \hline
     Model & ACER & APCER & BPCER  \\
    \hline
    Baseline & 9.0 & 18.1 & 0.0\\ 
    RGB Only & 5.7 & 10.0 & 1.4\\   
    DEP Only & 3.8 & 4.0 & 3.6\\    
    Feature Concat. & 1.1 & 1.5 & 0.7 \\ 
    Feature Concat. RGB & 1.3 & 2.1 & 0.6\\
    Feature Concat. DEP & 2.5 & 5.1 & \textbf{0.0}\\
    SA-Gate & 0.52 & \textbf{0.31} & 0.72\\
    \hline
    Ours & \textbf{0.27} & 0.39 & 0.14\\
    \hline
\end{tabular}
\caption{Ablation study in terms of features and modules on WMCA, GrandTest protocol.}
\label{Ablation}
\end{table}
\textit{LOO Protocol}: This evaluation aims to test the generalizability in detecting unseen attacks. As Table \ref{WMCA_LOO} shows, our model achieves the best overall performance and the individual scores are relatively high for most unseen spoof types.
Specifically, promising results are reached on Print, Replay, FakeHead and PaperMask. We also notice that, it does not behave so well for unseen spoofing glasses and silicon masks (FlexibleMask) although it is still leading by a large margin. The major reason for the failure on glasses is that the training attack signatures cover the entire faces, while their spoof traces only exist in local facial areas with most regions possessing a higher similarity with the  real faces. For silicon masks, their texture and geometric properties are inherently quite similar to the real faces, making them more challenging to detect. Overall, our model exhibits a plausible generalizability, which is also evidenced by the $t$-SNE visualization in Figure \ref{TNSE_LOO}. As we can see, the real and unseen attack samples reside in obviously distinct regions in the latent space. It confirms that distinguishable spoof traces are disentangled.

\begin{table}[t]
\centering
\begin{tabular}{c|c}
    \hline
     Loss & Mean ± Std (\%)  \\
    \hline
    w.o. $L_{center}$ & 12.4±13.6\\ 
    w.o. $L_{id}$ & 7.9±9.3\\
    w.o. $L_{intensity}$ & 6.8±7.3\\
    \hline
    Ours & \textbf{4.8±6.6}\\
    \hline
\end{tabular}
\caption{Ablation study in terms of losses on WMCA, LOO protocol.}
\label{Ablation_loss}

\end{table}

\subsubsection{CeFA}
This evaluation is conducted under Protocol 1, 2, and 4 (the cross-modality protocol is not applied). As Table \ref{table3} shows, the proposed model achieves competitive spoof detection results for all protocols by only using RGB-D data. In particular, 3D attacks are only included in the test set, which further demonstrates that our method has the favorable robustness and generalizability.

\subsection{Ablation Study}
Ablation is made on WMCA under the GrandTest protocol. Table \ref{Ablation} shows the results, where we compare the following: (1) \textbf{Baseline}: no disentanglement is performed. The classification network is trained and tested with the original RGB-D images; (2, 3) \textbf{RGB Only} \& \textbf{DEP Only}: The two branches of the disentanglement network are separately modeled without cross-modality feature fusion. The decision is made upon only one kind of spoof traces; (4, 5, 6) \textbf{Feature Concat.} \& \textbf{Feature Concat. RGB} \& \textbf{Feature Concat. DEP}: the cross-modality feature fusion is implemented by a vanilla concatenation, and spoof traces of different modalities are used for classification. (7) \textbf{SA-Gate}: our feature fusion module is replaced by the original version of SA-Gate.




We can observe: (i) Compared with Baseline, the disentanglement operation significantly improves the performance. Even if only one modality is used, better results are obtained (ACER 5.7\% of RGB Only \textit{v.s.} 3.8\% of DEP Only \textit{v.s.} 9.0\% of Baseline). (2) By comparing the entire model with \textbf{RGB Only} and \textbf{DEP Only}, the advantage of multi-modal data analysis is validated. A similar conclusion can be drawn from the comparison among (4, 5, 6); (iii) Compared with \textbf{Feature Concat.}, the proposed cross-modality fusion module improves ACER by 0.83\%, which highlights its contribution in reinforcing the correlations between two modalities. By capturing more complementary information, the disentanglement performance is promoted. (iv) Compared with \textbf{SA-gate}, our model improves ACER by 0.25\%, further validating that our fusion strategy is more effective.

In Table \ref{Ablation_loss}, we perform an ablation under the unseen protocol to verify the necessity of the \textbf{training losses}, where the mean ACER over all spoof types is reported. The successively decreasing ACER values demonstrate the effectiveness of the constraints.

\begin{figure}[t]
\centering
\includegraphics[width=0.41\textwidth]{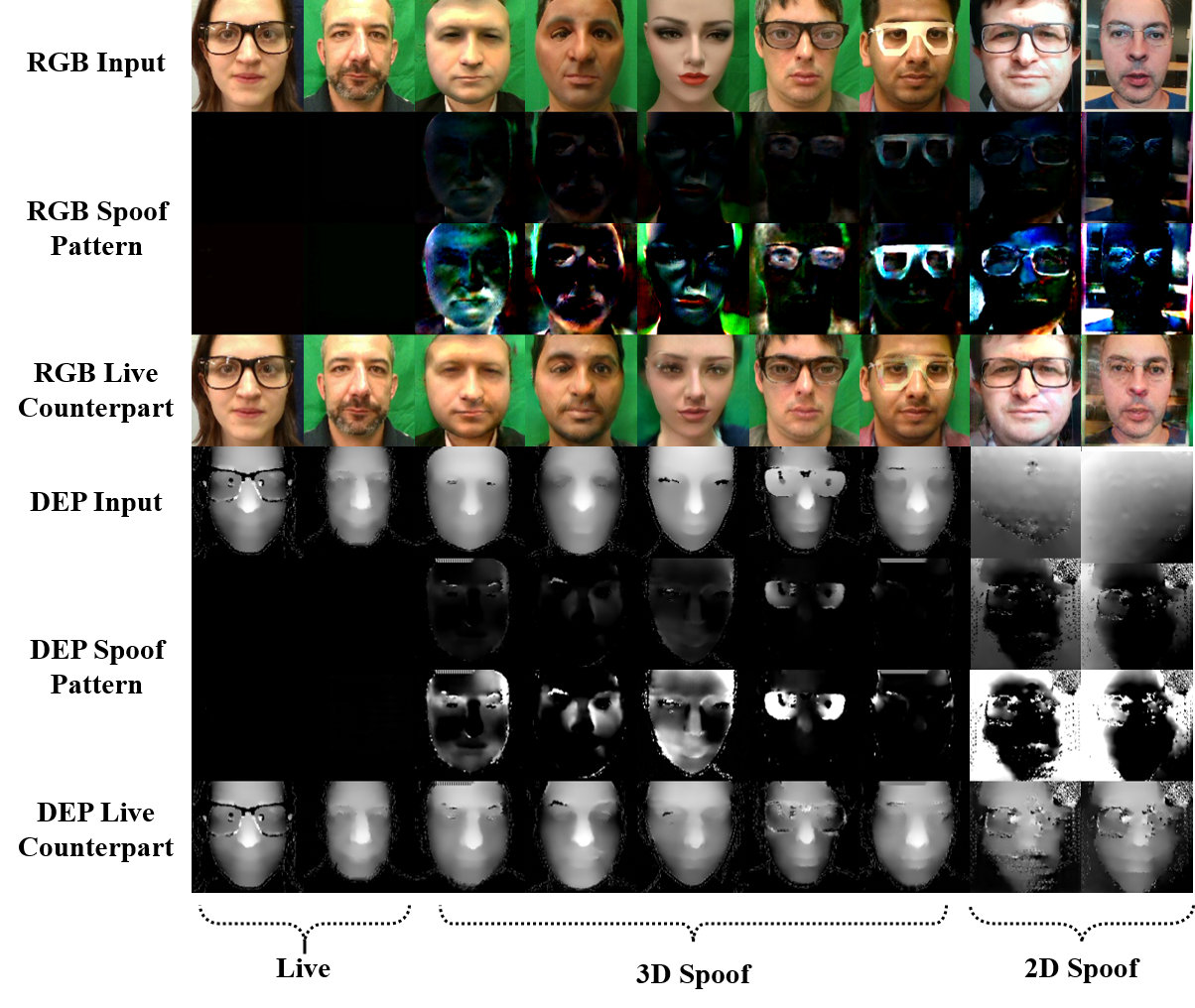} 
\caption{Visualization of the disentangled spoof traces on WMCA.}
\label{visualization_WMCA}

\end{figure}

\subsection{Spoof Trace Visualization}
The disentangled spoof traces and live counterparts are visualized in Figure \ref{visualization_WMCA}, with the original inputs shown in the 1st and 5th rows. The 2nd and 6th rows are the extracted spoof traces, which are re-normalized for better view, as displayed in the 3rd and 7th rows. As it can be seen, polysemantic spoof traces are successfully disentangled. Specifically, the RGB modality precisely captures the global color shift and unrealistic local patterns of 3D masks, \textit{e.g.}, shade of eyes, false boundaries, and texture differences (column 3-5). 
Even without pixel-wise supervision, the local differences around eyes can be detected (column 6-7). The spoof traces of 2D attacks are also well extracted, including the overexposure and color anomalies caused by photo recapturing (column 8-9). Comparatively, the depth modality focuses more on the anomalous geometric changes caused by spoofing, especially for 2D attacks.

\section{Conclusion}
This paper presents a novel multi-modal disentanglement network for FAS. By leveraging adversarial disentangled representation learning and cross-modality feature fusion, it generates complementary polysemantic spoof traces from RGB and depth images, respectively, contributing to detecting unseen attacks in more complex scenarios. In particular, it is demonstrated that multi-modal clues are beneficial to facilitating the disentanglement in each single modality. Extensive experiments are conducted on multiple databases and state-of-the-art results are reported.

\section{Acknowledgments}
This work is partly supported by the National Natural Science Foundation of China (No. 62176012, 62202031, 62022011), the Beijing Natural Science Foundation
(No. 4222049), the Research Program of State Key Laboratory of Software Development Environment (SKLSDE-2021ZX-04), and the Fundamental Research Funds for the Central Universities. We hereby give special thanks to Alibaba Group for their contribution to this paper.

\bibliography{aaai23}

\end{document}